\pdfoutput=1

\documentclass[11pt]{article}

\usepackage[preprint]{acl}

\usepackage{times}
\usepackage{latexsym}

\usepackage[T1]{fontenc}

\usepackage[utf8]{inputenc}

\usepackage{microtype}

\usepackage{inconsolata}

\usepackage{graphicx}
\usepackage{array}
\usepackage{multirow}
\usepackage{booktabs}
\usepackage{colortbl}
\usepackage{xcolor}
\usepackage{vcell}
\usepackage{arydshln}
\usepackage{amsmath}
\usepackage{amsfonts}
\usepackage{tabularx}
\usepackage{makecell}
\usepackage[normalem]{ulem}
\usepackage{CJKutf8}
\usepackage{pifont}
\usepackage{arydshln}
\usepackage{tikz}
\usepackage{pgfplots}
\usepackage{inconsolata}
\usepackage{fancyvrb}

\usepackage{todonotes}
\usepackage[capitalise,noabbrev]{cleveref}
\usepackage{listings}
\usepackage{fancyvrb} 
\makeatletter
\def\adl@drawiv#1#2#3{%
        \hskip.5\tabcolsep
        \xleaders#3{#2.5\@tempdimb #1{1}#2.5\@tempdimb}%
                #2\z@ plus1fil minus1fil\relax
        \hskip.5\tabcolsep}

\definecolor{verbgray}{gray}{0.9}

\usepackage{colortbl}
\definecolor{lightgray}{rgb}{0.7,0.7,0.7}

\newcommand{\cdashlinelr}[1]{%
  \noalign{\vskip 2pt}   
  \cdashline{#1}[.4pt/2pt] 
  \noalign{\vskip 2pt}   
}
\newcommand{\chinese}[1]{{\begin{CJK*}{UTF8}{gkai} #1 \end{CJK*}}}
\definecolor{light-orange}{HTML}{fee9d4}
\definecolor{light-green}{HTML}{d8f0d3}
\definecolor{light-blue}{HTML}{dae8f5}
\definecolor{light-red}{HTML}{FBC7C4}
\definecolor{set10-red}{HTML}{e41a1c}
\definecolor{set10-blue}{HTML}{377eb8}
\definecolor{set10-green}{HTML}{4daf4a}
\definecolor{bblue}{HTML}{4F81BD}
\definecolor{rred}{HTML}{c4260b}
\definecolor{ggreen}{HTML}{098c1f}
\definecolor{ppurple}{HTML}{9F4C7C}
\definecolor{oorange}{HTML}{F79646}

\usepackage{enumitem}
\setlist[itemize,enumerate]{leftmargin=*}
\usepackage[normalem]{ulem}
\usepackage{pgfplots}
\usetikzlibrary{pgfplots.groupplots}
\pgfplotsset{compat=1.3}
\usepackage{tikz}
\usetikzlibrary{patterns}
\usepackage{xcolor}
\usepackage{todonotes}
\usepackage{listings}
\usepackage{multirow}
\usepackage{graphicx}
\definecolor{CustomBlue}{RGB}{57,83,191}
\usepackage[most]{tcolorbox}
\definecolor{darkgreen}{HTML}{006400}

\newtcbox{\clustertab}[1]{on line, box align=base, colback={#1},colframe={#1},size=fbox,arc=2pt,top=-1.5pt, bottom=-1.5pt, left=-1.5pt, right=-1.5pt, boxrule=0pt, enlarge left by=1pt}

\title{MT-R1-Zero: Advancing LLM-based Machine Translation via R1-Zero-like Reinforcement Learning}

\makeatletter
\def\thanks#1{\protected@xdef\@thanks{\@thanks
        \protect\footnotetext{#1}}}
\makeatother

\author{
    Zhaopeng Feng$^{1}$\quad
    Shaosheng Cao$^{2\dag}$ \quad
    Jiahan Ren$^{1}$ \quad
    Jiayuan Su$^{1}$  \quad \\
    \bf Ruizhe Chen$^{1}$ \quad 
    \bf Yan Zhang$^{1}$ \quad
    \bf Zhe Xu$^{2}$ \quad 
    \bf Yao Hu$^{2}$ \quad
    \bf Jian Wu$^{1}$ \quad
    \bf Zuozhu Liu$^{1\dag}$ \\
    $^{1}$Zhejiang University \quad
    $^{2}$Xiaohongshu Inc. \quad \\
    \texttt{\{zhaopeng.23,zuozhuliu\}@intl.zju.edu.cn} \\
    \texttt{\{caoshaosheng,qiete,xiahou\}@xiaohongshu.com} \\
}

\thanks{$^{\dag}$ \space Corresponding author.}

\begin{document}
\maketitle


\begin{abstract}
Large-scale reinforcement learning (RL) methods have proven highly effective in enhancing the reasoning abilities of large language models (LLMs), particularly for tasks with verifiable solutions such as mathematics and coding.  However, applying this idea to machine translation (MT), where outputs are flexibly formatted and difficult to automatically evaluate with explicit rules, remains underexplored. In this work, we introduce \textbf{MT-R1-Zero}, the first open-source adaptation of the R1-Zero RL framework for MT without supervised fine-tuning or cold-start. We propose a rule-metric mixed reward mechanism to guide LLMs towards improved translation quality via emergent reasoning. On the WMT 24 English-Chinese benchmark, our MT-R1-Zero-3B-Mix achieves competitive performance, surpassing TowerInstruct-7B-v0.2 by an average of 1.26 points. Meanwhile, our MT-R1-Zero-7B-Mix attains a high average score of 62.25 across all metrics, placing it on par with advanced proprietary models such as GPT-4o and Claude-3.5-Sonnet, while the MT-R1-Zero-7B-Sem variant achieves state-of-the-art scores on semantic metrics. Moreover, our work exhibits strong generalization capabilities on out-of-distribution MT tasks, robustly supporting multilingual and low-resource settings. Extensive analysis of model behavior across different initializations and reward metrics offers pioneering insight into the critical role of reward design, LLM adaptability, training dynamics, and emergent reasoning patterns within the R1-Zero paradigm for MT. Our code is available at \href{https://github.com/fzp0424/MT-R1-Zero}{https://github.com/fzp0424/MT-R1-Zero}.
\end{abstract}

\begin{figure}[ht]
    \centering
    \includegraphics[scale=0.3]{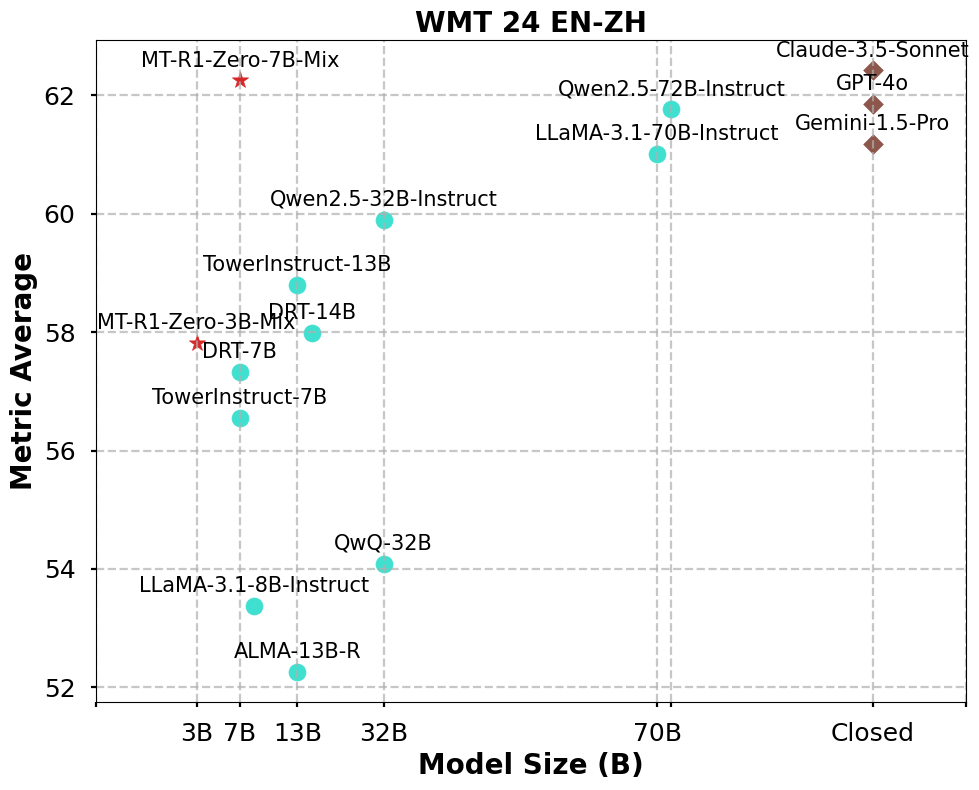}
    \caption{Performance comparison of contemporary LLM-based translation systems on the WMT 24 EN-ZH test set, plotted by average score across BLEU, COMETKiwi, and XCOMET versus model size.} 
    \vspace{-3mm}
    \label{fig:kl_plot}
\end{figure}

\section{Introduction}

Large-scale Reinforcement Learning (RL) has empowered Large Language Models (LLMs) with strong reasoning capabilities~\citep{o1, kimi, qwq32b}, demonstrating significant success in tasks such as mathematical reasoning or coding in which answers can be clearly verified. In particular, DeepSeek-R1-Zero~\citep{deepseekr1} introduced a pure rule-based RL approach that directly fosters emergent reasoning ability without requirements on structured Chain-of-Thought (CoT) data~\citep{wei2022chain,cui2025process} or sophisticated techniques such as Monte Carlo Tree Search (MCTS)~\citep{Silver2016MasteringTG, luo2024improve, qi2024mutual, guan2025rstar}. However, the applicability of these methods to machine translation (MT) remains challenging and underexplored, as MT outputs are flexibly generated and hard to evaluate automatically with explicit rules.

Recent work has launched attempts to empower LLMs for MT with reasoning capabilities~\cite{chen2025evaluating, liu2025new}. Early studies investigate explicit reasoning methods for improved translation, such as finetuning with CoT~\citep{drt-o1} or MCTS~\cite{marco-o1} , where advanced multi-step pipelines with self-correction or long-thought agentic mechanisms are further explored~\cite{feng2024improving,wang2024taste,drt-o1}. Another line of work leverages RL to empower LLMs for MT through process reward models or supervised finetuning (SFT) with manually annotated CoT data~\cite{feng2025mtreward,he2025r1t1}. However, these methods often depend on manually designed or synthetically generated structured CoT data, rely on complex search algorithms, or require explicit multi-stage prompting, leaving the potential of pure RL-based approaches largely unexplored. Furthermore, the performance reported in these studies often lags behind state-of-the-art (SoTA) open-source or proprietary models. 

Developing pure RL methods to directly enhance the reasoning ability of LLMs for better translation requires answering three key questions: 1) {\textbf{Feasibility}}: How to design R1-Zero-like RL pipelines with effective reward signals to directly solve MT tasks without binary rule-based rewards; 2) {\textbf{Reasoning capability}}: Could pure RL training cultivate emergent reasoning abilities and induce models to generate explicit thinking patterns for MT, such as multi-step CoT or verification/reflection; 3) {\textbf{Generalizability}}: Could the training paradigm generalizes across different models (e.g., pre-trained base models, instruction-tuned models, or models pre-trained on translation data) or diverse downstream settings (e.g., out-of-distribution, multilingual or low-resource scenarios).

In this work, we introduce \textbf{MT-R1-Zero}, the first open-source implementation that extends the RL-Zero-like RL training paradigm to MT. 
We propose a rule-metric mixed reward mechanism that adapts the original rule-based reward concept to effectively guide training in MT scenarios. We explore different rewards optimizing over lexical (Lex), semantic (Sem), and Lex-Sem mixed (Mix) objectives to guide LLMs towards improved translation quality via emergent reasoning. Our experiments demonstrate the efficacy of this approach: as RL training progresses, our MT-R1-Zero-3B-Mix achieves competitive performance, surpassing TowerInstruct-7B-v0.2 by an average of 1.26 points across all metrics (BLEU, COMETKiwi, XCOMET) on the WMT 24 English-Chinese (EN-ZH) benchmark. Meanwhile, our MT-R1-Zero-7B-Mix surpasses LLaMA-3.1-70B by an average of 1.24 points and Qwen2.5-72B by 0.48 points, even on par with top proprietary models such as GPT-4o and Claude-3.5-Sonnet. The MT-R1-Zero further demonstrate promising generalizability across multilingual and low-resource settings.  

Extensive experiments further provide key findings and insight into the adaptation of R1-Zero paradigm to MT. First, we empirically demonstrate that the choice of metric reward plays a pivotal role in steering RL optimization and translation style (semantic or lexical) (Finding 1). Further analysis reveals that MT-R1-Zero induces diverse emergent reasoning patterns, including dynamic language-of-thought transition during translation (Findings 2 and 3). We also identify distinct RL adaptability of different base LLMs (Finding 4). Ablation studies suggest that the pure RL process alone can lead to substantial translation improvements, independent of thinking verbosity (Section~\ref{sec:analysis}). Our core contributions are as follows: 

\begin{itemize}
\item We present the first open-source implementation of the DeepSeek-R1-Zero paradigm for MT, achieving superior performance across in-domain, OOD and generalization MT tasks.
\item Our analysis reveals key findings and recipes for effective R1-Zero adaptation to MT, including reward metric selection, emergent reasoning patterns, training dynamics and LLM adaptability.
\item Extensive experiments and ablations show that pure RL serves as the primary driver of MT improvements, with minimal dependence on forced reasoning or output length, highlighting the significant potential of RL for diverse translation applications and broader language tasks. 


\end{itemize}

\section{Related Work}
\noindent{\textbf{LLM Reasoning with Post-training.}}
Recent research indicates that scaling test-time computation can significantly enhance the ability of LLMs to tackle complex reasoning tasks~\citep{o1, zeng2024scaling, xiang2025towards}. Many approaches rely on sophisticated techniques such as step-level process reward models (PRMs) that provide granular feedback~\citep{lightman, yuan2024free, snell2024scaling} or MCTS to explore potential reasoning paths~\citep{feng2023alphazero, qi2024mutual, guan2025rstar}. A recent alternative, DeepSeek-R1-Zero~\citep{deepseekr1}, demonstrated that large-scale pure RL, guided only by formatting rules and correctness of final predictions (rule-based reward), can motivate LLMs to develop self-emergent reasoning processes for complex reasoning tasks. Subsequent work~\citep{OpenReasonerZero2025, openr1} successfully replicated this training paradigm in open-source models, focusing on mathematical domains. \citet{xie2025logic} further demonstrated the effectiveness and generalization capabilities of the R1-Zero paradigm using logic reasoning game problems, while ~\citet{huang2025vision} explored its potential for vision reasoning. Despite its potential, the application of the R1-Zero RL paradigm to complex generation tasks like MT, in which the accuracy/quality of outputs is not rule-based and difficult to validate automatically, remains an open question.

\noindent{\textbf{LLM Reasoning for MT.}}
Leveraging reasoning to improve MT has garnered increasing attention, as systematically explored in \citet{chen2025evaluating} and \citet{liu2025new}. Previous work have designed multi-step processes for MT, e.g., \citet{feng2024improving} introduced an API-based self-correcting framework, and \citet{wang2024taste} employed multi-task training followed by a multistage inference phase. \citet{drt-o1} integrated a similar procedure into inference-time CoT, using a multi-agent mechanism to synthesize long CoT prompts for English-Chinese literary translation. Efforts have also focused on reward modeling for MT reasoning. \citet{feng2025mtreward} constructed implicit process reward models for translation and explored their effectiveness when combined with test-time search. Recent study further evaluated explicit reasoning for MT using CoT fine-tuning and MCTS to expand test-time computation~\citep{marco-o1}. \citet{he2025r1t1} demonstrated that models can acquire reasoning-based translation capabilities through multi-stage training with manually constructed CoT templates.

However, these existing methods often necessitate manually designed or synthetically generated structured CoT data, rely on complex search algorithms (MCTS), or require explicit multi-stage prompting (self-correction). The effectiveness of large-scale pure RL training paradigms such as R1-Zero remains unexplored. 
Furthermore, the performance reported in these studies often lags behind state-of-the-art open-source or proprietary models. 


\section{Method}

In this section, we present our method that trains a translation model with pure RL using a hybrid reward model. Unlike tasks with fixed correct answers, translation allows for multiple valid outputs, making the evaluation more complicated. In this work, we introduce a rule-metric mixed reward that integrates reasoning format checking with multiple translation quality assessment metrics, which is used within the Group Relative Policy Optimization (GRPO) ~\citep{shao2024deepseekmath} algorithm to ensure stable and efficient RL training. 

\subsection{Rule-Metric Mixed Reward}
\label{sec:method_reward}

In RL, the reward is the main signal that drives model training. DeepSeek-R1-Zero \citep{deepseekr1} employs simple rule-based rewards that check whether the final answer is correct and whether the response follows a specific format. This works well for tasks with fixed format correct answers such as math or coding. However, there is often no single "correct" output for MT, impeding the design of rule-based rewards. Fortunately, the MT community has developed many evaluation metrics to measure translation quality.  Recent advancements in automated MT evaluation metrics have shown promise in aligning automated assessments with human translation quality judgments~\citep{freitag-etal-2022-results, freitag-etal-2023-results}. Thus, we design a rule-metric mixed reward, which consists of two parts: a Format Reward that checks output structure, and a Metric Reward that evaluates translation quality. We use a structured prompt template similar to that in DeepSeek-R1-Zero:

\begin{tcolorbox}[
    colframe=gray!80!black, 
    colback=gray!10!white, 
    coltitle=white, 
    fonttitle=\bfseries, 
    title=Template for MT-R1-Zero\label{long_open_q}, 
    boxrule=0.5mm, 
]

A conversation between User and Assistant. The User asks for a translation from \{src\_language\} to \{tgt\_language\}, and the Assistant solves it. The Assistant first thinks about the reasoning process in the mind and then provides the user with the final translation. The reasoning process and final translation are enclosed within <think> </think> and <translate> </translate> tags, respectively, i.e., <think> reasoning process here </think><translate> final translation here </translate>. \\
User:\{src\_text\} \\
Assistant:
\label{prompt}
\end{tcolorbox}
\noindent Here, \texttt{src\_language} and \texttt{tgt\_language} indicate the source and target languages, and \texttt{src\_text} denotes the source text requiring translation.

\paragraph{Format Reward:} We use regular expression extraction to enforce a structured response format. The model is required to place its reasoning process within \texttt{<think></think>} tags and provide the final translation inside \texttt{<translate></translate>} tags. The format reward score (\( S_{format} \)) is computed as:

\[
S_{format} =
\begin{cases}
\text{1}, & \text{if format is correct} \\
\text{-1}, & \text{if format is incorrect}
\end{cases}
\]

\paragraph{Metric Reward:} This reward evaluates the quality of model's translation, but \emph{only} if the response format is correct. We use automatic evaluation metrics to calculate a translation quality score \( S_{metric} \). We explore three approaches to compute \( S_{metric} \):

\begin{enumerate}
    \item \textbf{N-gram Lexical Matching Reward \textit{(Reward-Lex)}:} Metrics such as BLEU~\citep{papineni-etal-2002-bleu} or chrF~\citep{popovic-2015-chrf} evaluate translation quality by measuring the difference (primarily lexical overlap) between the translation and the human-written reference. In our experiments, we employ the BLEU score calculated via the \texttt{sacrebleu}\footnote{\href{https://github.com/mjpost/sacrebleu}{https://github.com/mjpost/sacrebleu}}.

    \item \textbf{Semantic and Contextual Reward \textit{(Reward-Sem)}:}Learning-based metrics like COMET~\citep{rei2020comet} and COMETKiwi~\citep{rei2022cometkiwi} are trained on human judgments (e.g., MQM quality assessments~\citep{freitag-etal-2021-experts}). These metrics can recognize good translations even if the wording differs from the reference, as long as the meaning is preserved. We use the COMETKiwi-23\footnote{\href{https://huggingface.co/Unbabel/wmt23-cometkiwi-da-xl}{https://huggingface.co/Unbabel/wmt23-cometkiwi-da-xl}}, which was used in the WMT 24~\citep{kocmi2024preliminary} and only needs the source sentence and the model’s translation.

    \item \textbf{Lexical and Semantic Mixed Reward \textit{(Reward-Mix)}:} To capture both lexical fidelity and semantic adequacy,  we use a hybrid reward \textit{(Reward-Mix)} that adds together Lexical Matching Reward \textit{(Reward-Lex)} and Semantic and Contextual Reward \textit{(Reward-Sem)}.

\end{enumerate}

\noindent Accordingly, the computation of \( S_{metric} \) depends on the selected reward configuration:
\[
\small
S_{metric} =
\begin{cases}
    \text{B}(\text{trans}, \text{ref}), & \text{if } \textit{Reward-Lex} \\
    \text{CK}(\text{src}, \text{trans}) & \text{if } \textit{Reward-Sem} \\
    \text{B}(\text{trans}, \text{ref}) + \text{CK}(\text{src}, \text{trans}), & \text{if } \textit{Reward-Mix}
\end{cases}
\]
where B denotes normalized BLEU score, CK denotes the COMETKiwi score, \texttt{trans} is the generated translation, \texttt{ref} is the reference translation, and \texttt{src} is the source text.


\paragraph{Rule-Metric Mixed Reward:}
The final reward \( r \) combines both the format reward (\( S_{format} \)) and the metric reward  (\( S_{metric} \)). Formally, it is calculated using the following rule:

\[
r =
\begin{cases}
S_{format} - 2, & \text{if } S_{format} = -1 \\
S_{format} + S_{metric}, & \text{if } S_{format} = 1 
\end{cases}
\]



where \( S_{metric} \) is calculated only  if the response format is correct \( S_{format} = 1 \). If the format is incorrect (\( S_{format} = -1 \)), we skip the metric reward evaluation and assign a fixed penalty (e.g., 2) to discourage format violations. This setup encourages the model to first learn the correct output structure. When the format is correct, the final reward becomes \( r = 1 + S_{metric} \). Unlike traditional rule-based rewards that give a fixed score for correct outputs, our approach uses a continuous metric score. This means the reward can vary within the [1, 2] or [1, 3] range, depending on translation quality. As a result, the model receives more detailed feedback and can learn to improve even small differences in translation quality across correctly formatted outputs.

\subsection{RL Algorithm}

We use the Group Relative Policy Optimization (GRPO) algorithm~\citep{shao2024deepseekmath} to train the translation model with our rule-metric mixed reward. In each training step, for a given translational question $q$, we sample a group of candidate outputs $\{o_1,o_2,\cdots,o_G\}$ from the policy model $\pi_{\theta_{old}}$. $A_i = \frac{r_i - \operatorname{mean}(\{r_1, r_2, \dots, r_G\})}{\operatorname{std}(\{r_1, r_2, \dots, r_G\})}$ is the computed advantage using the group rule-metric mixed rewards $\{r_1,r_2,\cdots,r_G\}$. GRPO then maximizes the following objective function to optimize $\pi_{\theta}$:  
\begin{equation}
\begin{aligned}
J_{\mathrm{GRPO}}(\theta) 
&= \mathbb{E}_{q \sim P(Q),\, \{o_i\}_{i=1}^G \sim \pi_{\theta_{\mathrm{old}}}(O \mid q)} \\
&\Biggl[
  \frac{1}{G} \sum_{i=1}^G
  \min\!\Bigl(
    \frac{\pi_{\theta}(o_i \mid q)}{\pi_{\theta_{\mathrm{old}}}(o_i \mid q)}\,A_i,\, \\
    &\mathrm{clip}\!\Bigl(
      \frac{\pi_{\theta}(o_i \mid q)}{\pi_{\theta_{\mathrm{old}}}(o_i \mid q)},
      1-\varepsilon,\,
      1+\varepsilon
    \Bigr)
    A_i
  \Bigr)  \\
  &-\,\beta\,D_{\mathrm{KL}}\bigl(\pi_{\theta}\,\big\|\,\pi_{\mathrm{ref}}\bigr)
\Biggr],
\end{aligned}
\label{eq1}
\end{equation}
where $\varepsilon$ and $\beta$ are hyperparameters controlling the PPO clipping threshold and the weight of the Kullback–Leibler (KL) divergence penalty~\cite{schulman2017proximal,shao2024deepseekmath}, respectively. Specifically, $\varepsilon$ determines the permissible range for policy updates, while $\beta$ regulates the magnitude of the KL penalty during training to prevent excessive policy shifts from the reference policy $\pi_{ref}$ (typically the initialization of $\pi_{\theta}$). $D_{KL}\bigl(\pi_\theta \,\|\, \pi_{\text{ref}}\bigr)
= \frac{\pi_{\text{ref}}(o_i \mid q)}{\pi_\theta(o_i \mid q)}
- \log\!\Bigl(\frac{\pi_{\text{ref}}(o_i \mid q)}{\pi_\theta(o_i \mid q)}\Bigr)
- 1 \,$ is the KL divergence approximation term.

\section{Experiments}
\subsection{Experimental Setup}
\label{sec:exp_set}

\noindent{\textbf{Dataset and Benchmarks.}}
Our primary experimental focus is on English (EN) and Chinese (ZH). Following \citet{xu2023paradigm} and \citet{feng-etal-2024-ladder}, we collect parallel examples (EN$\rightleftharpoons$ZH) sourced from WMT 2017 through WMT 2020. We apply a filter to exclude sentences containing fewer than 30 characters, leading to a final training set of 13,130 examples. For evaluation, we assess performance on two in-domain translation tasks using recent WMT benchmarks: EN-ZH (WMT 24\footnote{\href{https://www2.statmt.org/wmt24/translation-task.html}{https://www2.statmt.org/wmt24/translation-task.html}}) and ZH-EN (WMT 23\footnote{\href{https://www2.statmt.org/wmt23/translation-task.html}{https://www2.statmt.org/wmt23/translation-task.html}}). Additionally, we evaluate generalization capabilities on three out-of-distribution (OOD) translation directions: English-Japanese (EN-JA, WMT 2024), German-English (DE-EN, WMT 2023 Document-level), and German-Chinese (DE-ZH, Flores-200~\citep{costa2022no}). Detailed statistics are presented in Table~\ref{tab:main_data}. 

\noindent{\textbf{Baselines.}}
Our primary baselines encompass leading proprietary models, namely Claude-3.5-Sonnet~\citep{claude35}, GPT-4o~\citep{openai2023gpt4}, and Gemini-1.5-Pro~\citep{team2024gemini15}, alongside advanced open-source models such as the Qwen2.5 series~\citep{yang2024qwen25}, LLaMA-3.1 series~\citep{grattafiori2024llama}, and the translation-specific Tower family~\citep{alves2024tower}. Proprietary models were accessed via their APIs\footnote{The specific proprietary models accessed include Anthropic's \texttt{claude-3-5-sonnet-20241022}, OpenAI's \texttt{gpt-4o-2024-08-06}, and Google's \texttt{gemini-1.5-pro}.}. More evaluation details can be found in Appendix~\ref{app:inference}.

\noindent{\textbf{Evaluation Metrics.}}
We assess translation quality using a suite of three complementary metrics: the lexical metric BLEU~\citep{post-2018-call}, the reference-free learning-based metric COMETKiwi~\citep{rei2022cometkiwi} (COMETKiwi-23-XL), and the reference-based learning-based metric XCOMET~\citep{guerreiro2024xcomet} (XCOMET-XL). Together, these metrics provide a comprehensive view by evaluating both lexical fidelity and semantic adequacy.

\noindent{\textbf{Training Details.}} 
Our implementation is based on the verl\footnote{\href{https://github.com/volcengine/verl}{https://github.com/volcengine/verl}} framework. We selected the Qwen2.5-base series (3B and 7B parameter variants) as starting models for MT-R1-Zero training. During training, we configure a batch size of 8 and utilize 8 rollouts per prompt within the GRPO algorithm. We employ a constant learning rate of 5e-7 and set the sampling temperature to 1.0. The maximum generation length for responses is capped at 1024 tokens.  We set the KL penalty coefficient 
$\beta$ to 0, thereby removing the KL constraint against the reference policy. This decision stems from our empirical observation that the KL penalty tends to restrict the model's exploration of diverse response lengths, which we will discuss further in Section~\ref{sec:kl}. The PPO clipping range $\epsilon$ is set to 0.2. All models are trained for 1 epoch on 4 NVIDIA H800 80G GPUs for about 13 hours. 

\begin{table*}[t]
    \centering
    \small
    \resizebox{\textwidth}{!}{%
    \setlength{\tabcolsep}{4pt}
    \begin{tabular}{l*{9}{c}} 
        \toprule
        \multirow{2.5}{*}{\sc Model} & \multicolumn{4}{c}{\sc ZH-EN} & & \multicolumn{4}{c}{\sc EN-ZH} \\ 
        \cmidrule(lr){2-5} \cmidrule(lr){7-10} 
         & BLEU & COMETKiwi & XCOMET & Avg. & & BLEU & COMETKiwi & XCOMET & Avg. \\ 
        \midrule
        \multicolumn{10}{c}{\textit{\textbf{Closed}}} \\ 
        Claude-3.5-Sonnet (2024/10)     & 22.55 & 71.69 & 87.32 & 60.52 & & 38.63 & 70.39 & 78.24 & 62.42 \\ 
        GPT-4o (2024/08)                  & 22.57 & 71.63 & 87.22 & 60.47 & & 41.13 & 69.01 & 75.43 & 61.86 \\ 
        Gemini-1.5-Pro (2025/03)  & 18.34 & 69.23 & 85.55 & 57.71 & & 39.82 & 67.47 & 76.26 & 61.18 \\
        \midrule

        \multicolumn{10}{c}{\textit{\textbf{Open}}} \\ 
        \multicolumn{10}{@{}l}{\textcolor{lightgray}{\textit{General Purpose LLMs}}} \\
        LLaMA-3.1-70B-Instruct   & 25.19 & 70.43 & 86.21 & 60.61 & & 39.82 & 68.05 & 75.17 & 61.01 \\ 
        Qwen2.5-72B-Instruct    & 21.96 & 70.95 & 87.07 & 59.99 & & 39.29 & 69.04 & 76.97 & 61.77 \\
        Qwen2.5-32B-Instruct    & 20.54 & 69.35 & 85.47 & 58.45 & & 36.36 & 68.43 & 74.90 & 59.90 \\

        \multicolumn{10}{@{}l}{\textcolor{lightgray}{\textit{Translation-Specific LLMs}}} \\
        TowerInstruct-13B-v0.1  & 24.72 & 70.17 & 85.69 & 60.19 & & 37.06 & 66.22 & 73.13 & 58.80 \\
        TowerInstruct-7B-v0.2   & 23.32 & 69.99 & 84.93 & 59.41 & & 34.93 & 64.04 & 70.67 & 56.55 \\
        \midrule

        \multicolumn{10}{c}{\textit{\textbf{Ours}}} \\ 
        Qwen2.5-3B-Base      & 14.26 & 64.86 & 76.76 & 51.96 & & 15.90 & 52.05 & 67.13 & 45.03 \\
        MT-R1-Zero-3B-Lex      & 21.53 & 66.33 & 81.69 & 56.52 & & 33.70 & 60.58 & 65.67 & 53.32 \\
        MT-R1-Zero-3B-Sem & 18.41 & 70.33 & 85.98 & 58.24 & & 24.32 & 69.75 & 76.92 & 57.00 \\
        MT-R1-Zero-3B-Mix & 22.54 & 68.84 & 84.08 & 58.49 & & 36.27 & 65.05 & 72.10 & 57.81 \\

        \cdashlinelr{1-10}
        Qwen2.5-7B-Base      & 18.23 & 68.27 & 84.99 & 57.16 & & 31.14 & 63.38 & 69.83 & 54.78 \\
        MT-R1-Zero-7B-Lex      & 23.56 & 65.35 & 82.12 & 57.01 & & 40.11 & 64.57 & 70.21 & 58.30 \\
        MT-R1-Zero-7B-Sem & 16.62 & 71.66 & 86.07 & 58.12 & & 23.07 & 72.07 & 79.37 & 58.17 \\
        MT-R1-Zero-7B-Mix & 23.98 & 70.81 & 86.17 & 60.32 & & 40.97 & 69.43 & 76.36 & 62.25 \\

        \bottomrule
    \end{tabular}
    }
    \caption{
    Performance comparison on in-domain translation directions (EN-ZH, ZH-EN) using BLEU, COMETKiwi, and XCOMET metrics, with average metric scores (Avg.). MT-R1-Zero variants (\textit{-Lex}, \textit{-Sem}, \textit{-Mix}) are compared against closed and open baselines, which are further categorized by accessibility and specialization. The \textit{-Mix} variant often achieves the best balance, while \textit{-Sem} reaches peak semantic scores. 
    }
    \label{tab:main_mt_results}
\end{table*}

\begin{table}[htbp]
    \centering
    \small
    \resizebox{\columnwidth}{!}{%
    \setlength{\tabcolsep}{5pt}
    \begin{tabular}{@{}lcccc}
        \toprule
        \multirow{2.5}{*}{\sc Model} & \multicolumn{4}{c}{\sc Out-of-distribution} \\
        \cmidrule(lr){2-5}
                                      & \sc EN-JA & \sc DE-EN (Doc) & \sc DE-ZH & Avg. \\
        \midrule
        \multicolumn{5}{@{}l}{\textcolor{lightgray}{\textit{Strong Baseline}}} \\
        Qwen2.5-72B-Instruct           & 76.86          & 89.51                & 88.42 & 84.93 \\
        LLaMA3.1-70B-Instruct          & 75.67          & 88.72                & 87.42 & 83.94 \\
        \multicolumn{5}{@{}l}{\textcolor{lightgray}{\textit{Same-size Baseline}}} \\
        Qwen2.5-7B-Instruct            & 63.74          & 87.45                & 84.43 & 78.54 \\
        LLaMA-3.1-8B-Instruct          & 64.50          & 86.84                & 82.23 & 77.86 \\
        TowerInstruct-7B-v0.2          & 56.73          & 89.47                & 84.28 & 76.83 \\
        \midrule
        MT-R1-Zero-7B-Lex                      & 60.65          & 85.25                & 83.86 & 76.59 \\
        MT-R1-Zero-7B-Sem                       & 71.95          & 87.68                & 87.66 & 82.43 \\
        MT-R1-Zero-7B-Mix                  & 68.49          & 88.69                & 88.69 & 81.96 \\
        \bottomrule
    \end{tabular}
    }
    \caption{
    Out-of-distribution performance comparison using the XCOMET metric on EN-JA, DE-EN (Document-level), and DE-ZH. 
    }
    \vspace{-1mm}
    \label{tab:ood_xcomet}
\end{table}

\subsection{Main Results}
\noindent{\textbf{In-Domain Performance.}} 
Our models show substantial gains over their corresponding base versions,
and exhibit competing performance compared to existing SoTA benchmarks (Table~\ref{tab:main_mt_results}). On the EN-ZH direction, our MT-R1-Zero-7B-Mix on the average score (62.25) also surpasses GPT-4o (61.86) and Qwen2.5-72B (61.77). In addition, the MT-R1-Zero-7B-Sem achieves the best semantic-level performance on EN-ZH, scoring 72.07 on COMETKiwi and 79.37 on XCOMET. This surpasses the strongest proprietary model, Claude-3.5-Sonnet, by 1.68 COMETKiwi points and exceeds the best listed open-source model, Qwen2.5-72B, by more than 3 points.  On the ZH-EN direction, MT-R1-Zero-7B-Mix is also highly competitive. Our MT-R1-Zero-7B-Sem achieves a COMETKiwi score of 71.66, which is comparable to the top closed models (Claude-3.5-Sonnet 71.69, GPT-4o 71.63) and surpasses strong open-source models such as LLaMA-3.1-70B (70.43) and Qwen2.5-72B (70.95). 
Furthermore, the MT-R1-Zero-3B-Sem delivers impressive performance for its scale. It scores 69.75 COMETKiwi on EN-ZH, which is approximately 1.7 points higher than the much larger LLaMA-3.1-70B and over 0.7 points above Qwen2.5-72B. 


\noindent{\textbf{Out-of-Distribution Performance.}} 
Table~\ref{tab:ood_xcomet} reports the XCOMET of our models on OOD language pairs with a zero-shot setting (models trained only on EN-ZH/ZH-EN). Despite this challenging setup, our models exhibit strong generalization. The MT-R1-Zero-7B-Sem achieves the highest average XCOMET score (82.43) across the OOD tasks, reaching top scores on EN-JA (71.95) and DE-EN (87.68). The MT-R1-Zero-7B-Mix also demonstrates highly competitive generalization with an average score of 81.96, and secures the highest score on DE-ZH (88.69). While these variants do not consistently surpass the much larger strong baselines (Qwen2.5-72B Avg. 84.93, LLaMA3.1-70B Avg. 83.94), they are still highly competitive. Crucially, MT-R1-Zero-7B-Sem and -Mix significantly outperform all same-size baselines (Qwen2.5-7B-Instruct Avg. 78.54, LLaMA-3.1-8B-Instruct Avg. 77.86, TowerInstruct-7B-v0.2 Avg. 76.83) by a considerable margin (at least 3.4 points). These OOD results suggest that the quality improvements in MT-R1-Zero can effectively transfer to unseen language pairs.
Results using COMETKiwi and BLEU are also provided in Appendix Tables~\ref{tab:ood_cometkiwi} and \ref{tab:ood_bleu}, respectively.

\begin{figure*}[ht]
    \centering
    \includegraphics[scale=0.31]{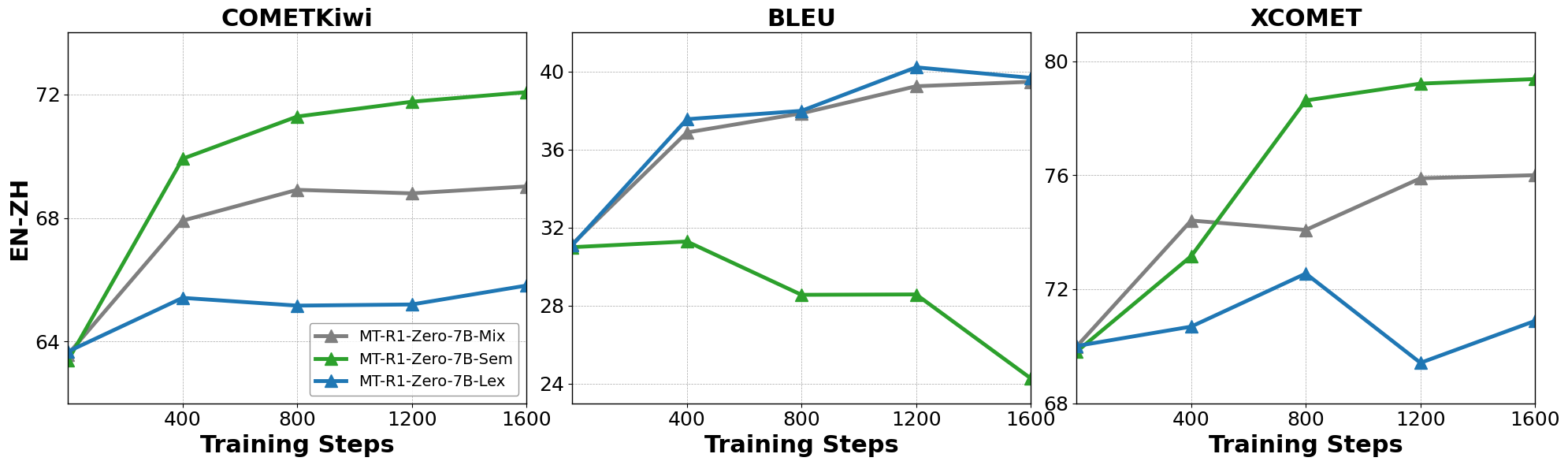}
    \caption{Training dynamics using \textit{Reward-Lex}, \textit{Reward-Sem}, and \textit{Reward-Mix}, evaluated with COMETKiwi, BLEU, and XCOMET. 
    } 
    \vspace{-3mm}
    \label{fig:merge_bleu_comet}
\end{figure*}

\begin{figure*}[ht]
    \centering
    \includegraphics[scale=0.75]{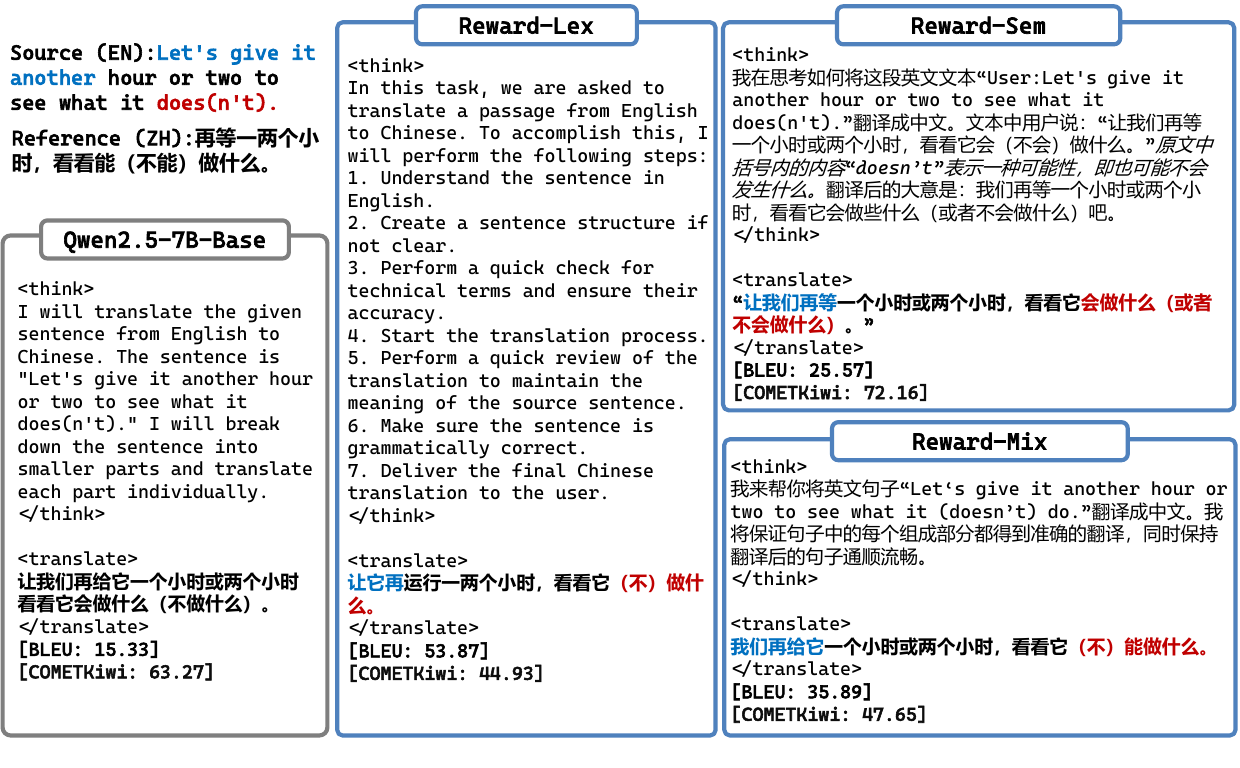}
    \caption{Qualitative examples illustrates the effect of different reward functions (\textit{Reward-Lex}, \textit{Reward-Sem}, \textit{Reward-Mix}) on EN-ZH translation, where the stylistic differences are driven by reward optimization (Finding 1).} 
    \vspace{-3mm}
    \label{fig:case_metric}
\end{figure*}

\begin{figure*}[ht]
    \centering
    \includegraphics[scale=0.29]{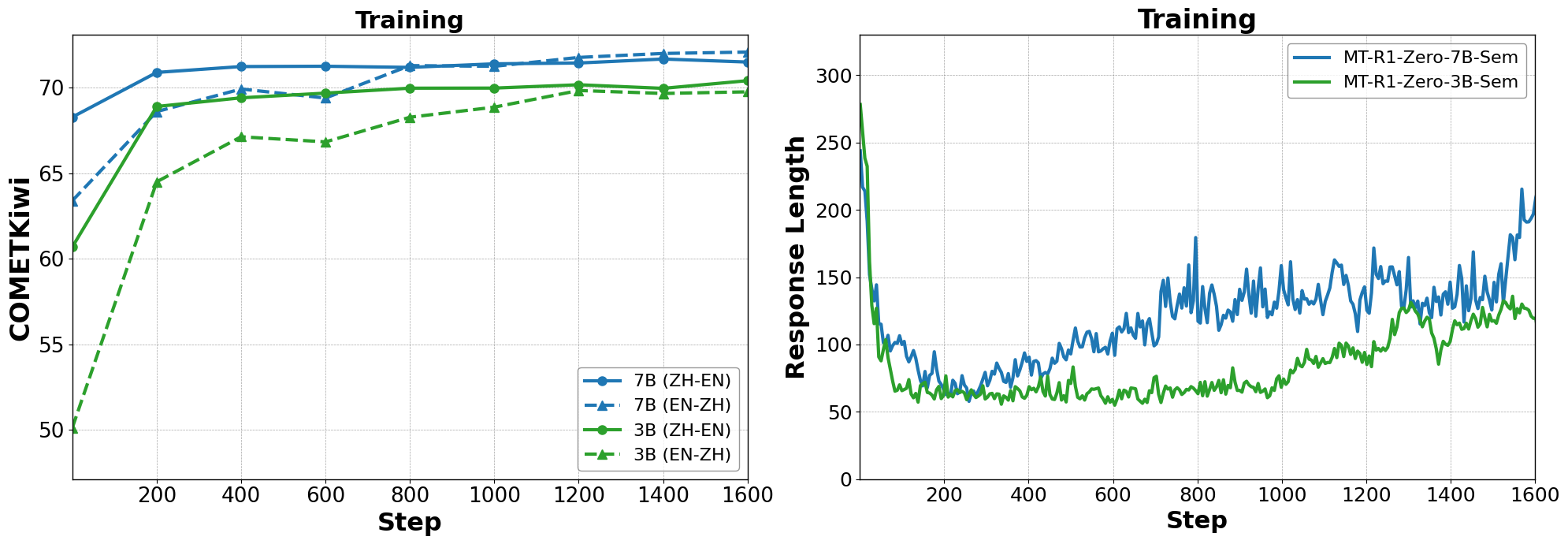}
    \caption{Training dynamics of MT-R1-Zero models (using Reward-Sem). \textbf{Left}: COMETKiwi score progression for 3B and 7B models on EN-ZH and ZH-EN test sets. \textbf{Right}: Average response length changes over training steps, exhibiting the classic decrease-then-increase pattern (Finding 2).} 
    \vspace{-3mm}
    \label{fig:main_fig}
\end{figure*}

\begin{figure*}[ht]
    \centering
    \includegraphics[scale=0.72]{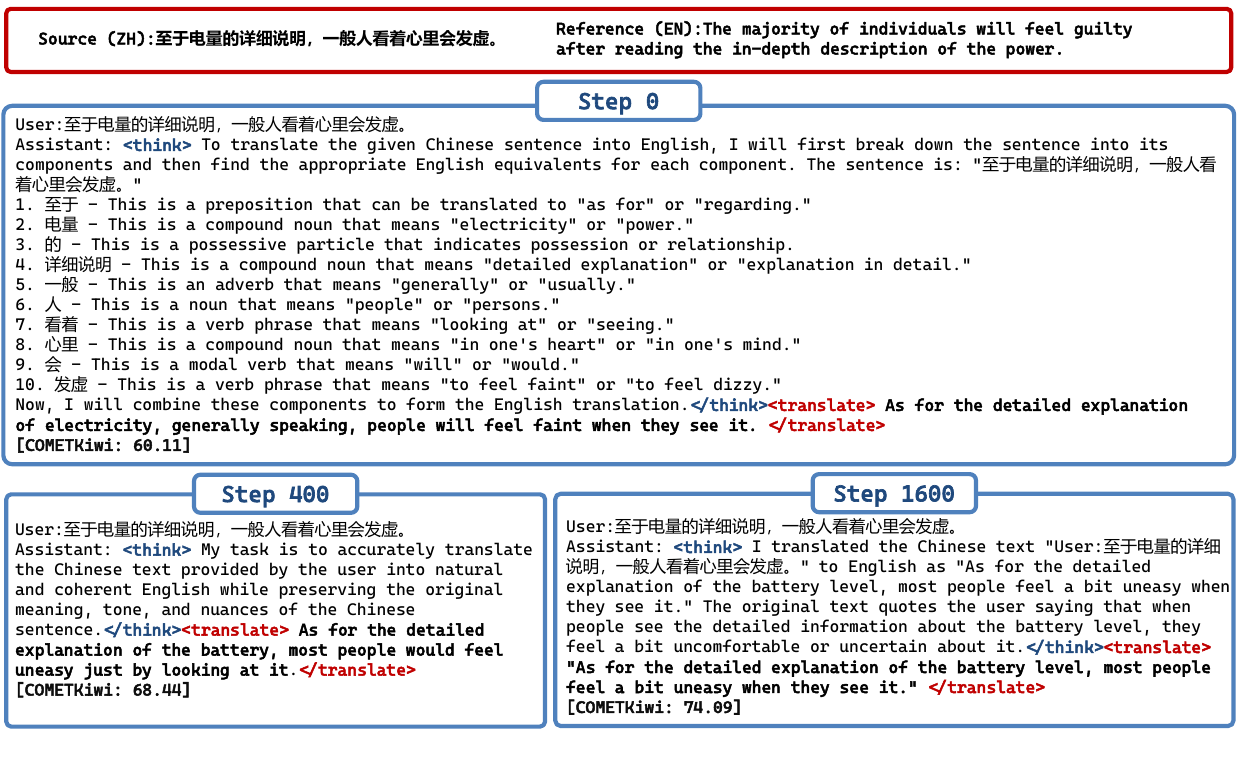}
    \caption{Evolution of an MT-R1-Zero model's reasoning process and translation output for the Chinese source text \chinese{"其影响可能类似于2008年的经济危机"} at different training steps (0, 400, 1600), showcasing the shift from decomposition to more semantic analysis (Finding 2). } 
    \vspace{-3mm}
    \label{fig:case_main}
\end{figure*}

\begin{figure*}[ht]
    \centering
    \includegraphics[scale=0.76]{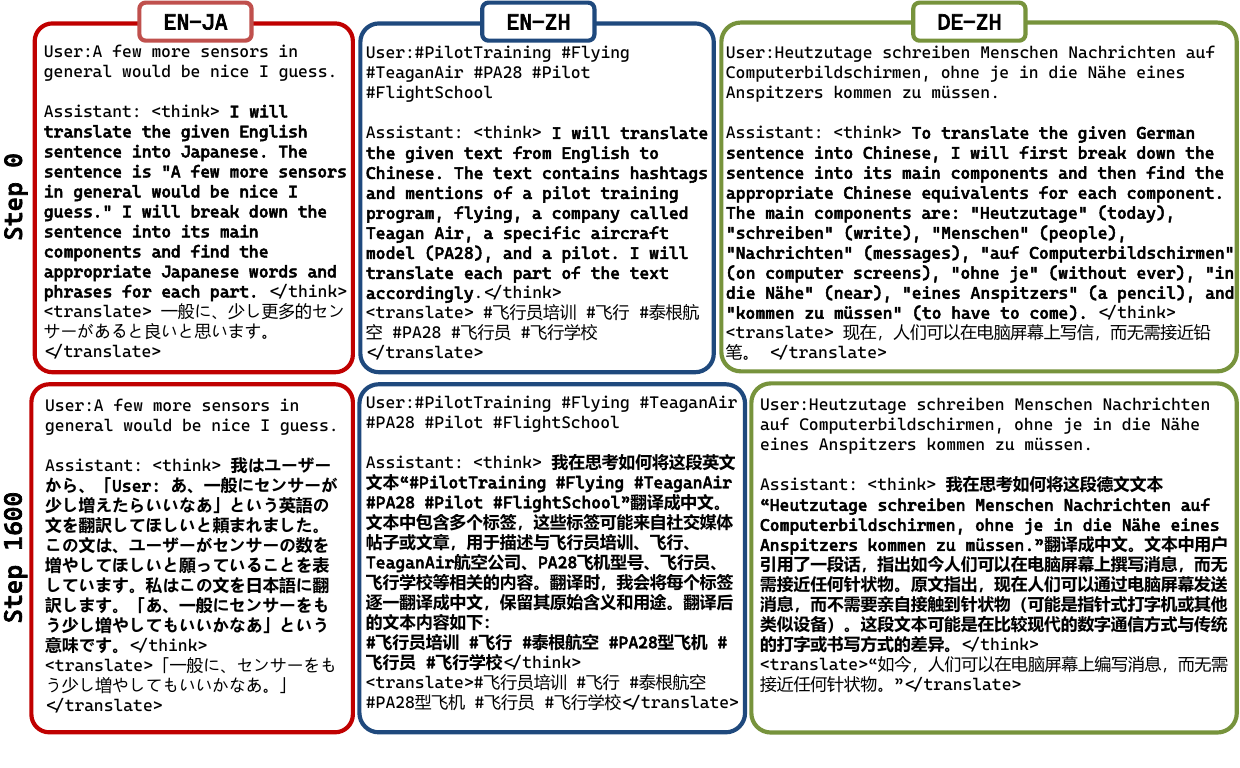}
    \caption{Examples illustrating language-of-thought phenomenon, i.e., transition of the internal reasoning language in MT-R1-Zero models. The reasoning language transits from English at Step 0 to target language at Step 1600, indicated by \textbf{bold} text across various OOD test pairs (Finding 3).} 
    \vspace{-3mm}
    \label{fig:case_multi_language}
\end{figure*}

\section{Key Findings and Insight}
\label{sec:key_findings}
Based on our extensive experiments adapting the R1-Zero paradigm to MT, we identify several key findings regarding the underlying mechanisms, design ideas, and emergent behaviors of our MT-R1-Zero framework.

\subsection{Impact of Reward Metric Selection}
\label{sec:reward_select}
As detailed in Section~\ref{sec:method_reward}, we explore three metric rewards: \textit{Reward-Lex}, \textit{Reward-Sem}, and \textit{Reward-Mix}. Our results demonstrate that the choice among these significantly affects the learning target and final model outputs, as stated in Finding 1. 
\begin{tcolorbox}[
    colback=gray!10!white, 
    colframe=darkgreen,
    boxrule=0.5mm, 
]
\textbf{Finding 1:} Reward metric selection critically shapes optimization targets and translation style. 
\end{tcolorbox}
Figure~\ref{fig:merge_bleu_comet} presents the training dynamics with different rewards. Training with \textit{Reward-Lex} maximizes BLEU scores, often at the expense of semantic scores, while \textit{Reward-Sem} maximizes COMETKiwi, leading to a decline in BLEU. Training with \textit{Reward-Mix} improves both metrics, with a trade-off of achieving sub-optimal COMETKiwi compared to \textit{Reward-Sem}. Independent evaluation with XCOMET further supports this finding, showing consistent improvements for Sem and Mix variants while fluctuating for Lex. This finding aligns with the insight from \citet{chen2025evaluating}, suggesting that lexical and semantic assessments are complementary, particularly for reasoning-oriented LLMs, and combining them can offer a more comprehensive evaluation signal.

Qualitatively (Figure~\ref{fig:case_metric}), this optimization alignment manifests as distinct translation styles. BLEU optimization encourages literal, n-gram focused translations, potentially sacrificing nuance. COMETKiwi optimization fosters translations that prioritize semantic faithfulness, even if lexically divergent from references. In contrast, the mixed reward yields balanced translations. This demonstrates that the metric reward fundamentally dictates the nature of the translation quality learned (e.g., semantic v.s. lexical). Therefore, careful metric selection and deliberate fusion are essential for tailoring RL-based MT refinement towards specific and desired translations.

\subsection{Emergence and Evolution of Translation Thinking Patterns}
By observing the training process, we provide several insights into model adaptation and the emergence of reasoning.
\begin{tcolorbox}[
    colback=gray!10!white, 
    colframe=darkgreen, 
    boxrule=0.5mm, 
]
\textbf{Finding 2:} Response length initially declines rapidly and then gradually increases as training progresses.
\end{tcolorbox}
Figure~\ref{fig:main_fig} (Right) depicts the pattern in \textit{Finding 2} alongside consistent COMETKiwi improvements (Left). Qualitative analysis (Figure~\ref{fig:case_main}) reveals that this length trajectory reflects evolving reasoning strategies. The initial decline corresponds to the model mastering the required format while transitioning from naive decomposition (Step 0) to more efficient, direct translations. The subsequent increase aligns with the development of richer semantic analysis and deeper contextual reasoning within the \texttt{<think></think>} tags (Step 1600).
\begin{tcolorbox}[
    colback=gray!10!white, 
    colframe=darkgreen, 
    boxrule=0.5mm, 
]
\textbf{Finding 3:} Diverse reasoning patterns emerge autonomously, varying in style and complexity, and moreover, the internal reasoning language could dynamically transit to target languages even for OOD settings. 
\end{tcolorbox}

As R1-Zero-like training typically lacks a cold-start~\citep{deepseekr1,huang2025vision} phase with predefined reasoning examples, the observed thinking processes should be emergent and shaped by the RL objective. Our framework incentivizes a variety of reasoning styles within the \texttt{<think></think>} tags (Figure~\ref{fig:case_kl_long_think}). In particular, we observe patterns ranging from structured multi-step decomposition (Types I-III) to more colloquial processing (Types IV-V). While some instances include explicit "review/refine" steps, these generally appear as pre-planned components rather than the conversational, iterative self-correction characteristic of the "Aha moment" reported in mathematical reasoning tasks~\citep{deepseekr1,xie2025logic, OpenReasonerZero2025}. This suggests that while MT-R1-Zero successfully encourages thinking, the complexity and specific nature of emergent reasoning are task-dependent. 

Furthermore, we observe a striking and interesting \textbf{"language-of-thought" (transition in the language used for internal reasoning}) phenomenon during OOD testing (Figure~\ref{fig:case_multi_language}). While base models often use English as default thinking language based on template, MT-R1-Zero models progressively transit to utilize the \textbf{target language} of the translation task for their reasoning process within the \texttt{<think></think>} block during training (see bold Japanese or Chinese text in step 1600). This dynamic adaptation of the internal "language of thought", conditioned on the task, emerges even without direct supervision on reasoning language.

\begin{figure*}[ht]
    \centering
    \includegraphics[scale=0.31]{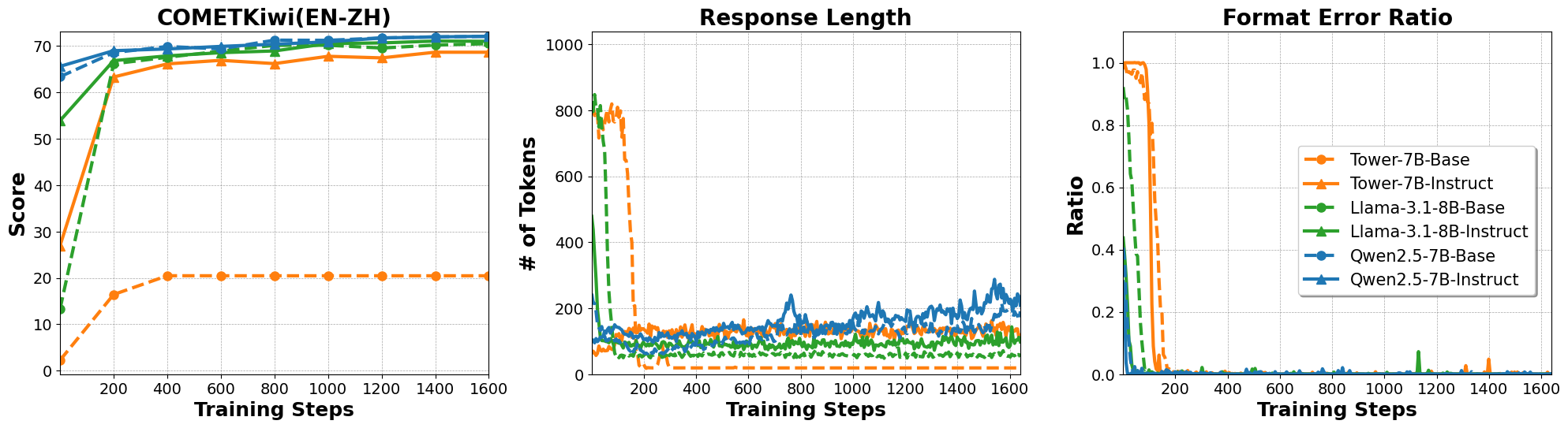}
    \caption{Comparison of training dynamics for different model families (Qwen2.5, LLaMA-3.1, Tower) undergoing MT-R1-Zero RL training, highlighting differences in adaptability (Finding 4).} 
    \vspace{-3mm}
    \label{fig:train_dynamic_multi}
\end{figure*}

\begin{figure*}[ht]
    \centering
    \includegraphics[scale=0.75]{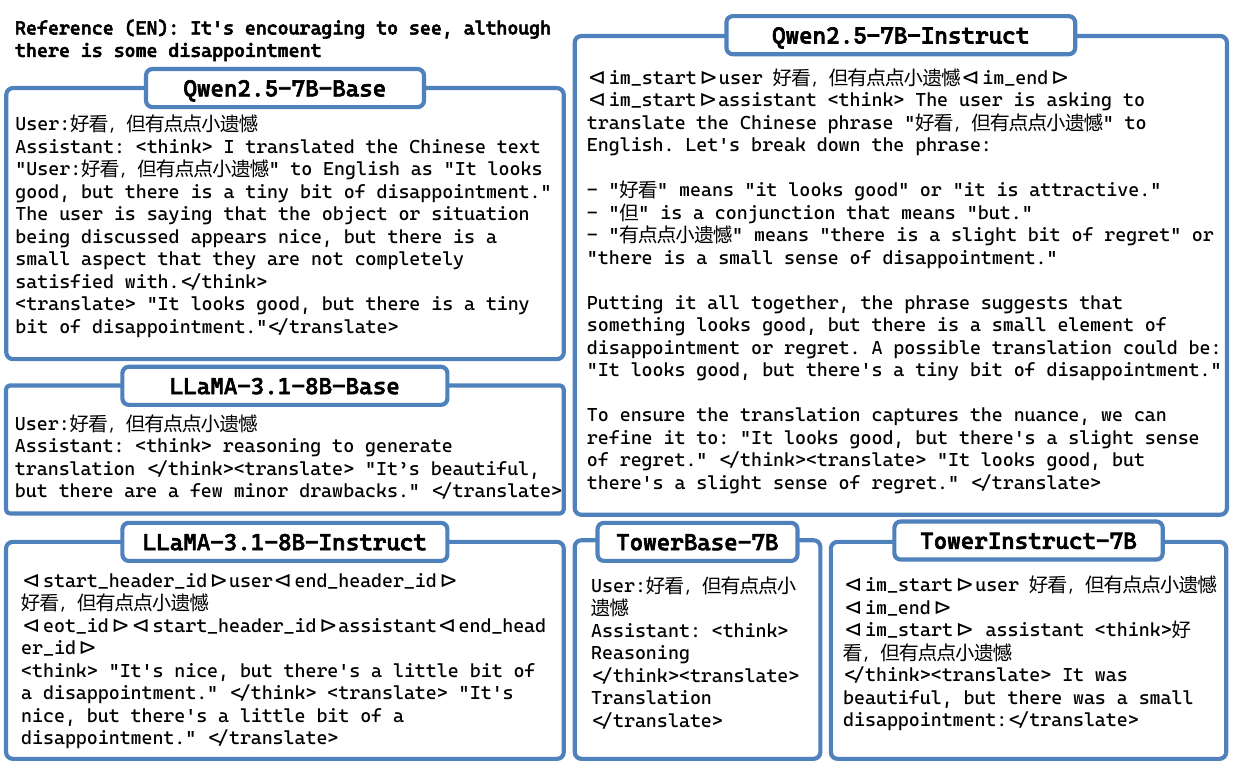}
    \caption{Qualitative comparison of final outputs from different starting models trained with MT-R1-Zero for the Chinese input \chinese{"好看,但有点点小遗憾"}, illustrating varying degrees of format adherence and reasoning generation, including format hacking by some models (Finding 4).} 
    \vspace{-3mm}
    \label{fig:case_multi_model}
\end{figure*}

\subsection{Training Dynamics of Different LLMs}
The effectiveness and training behavior of MT-R1-Zero are significantly influenced by the base LLM architecture and its initial state (pre-trained vs. instruction-tuned). We compare models from three distinct families: general-purpose (Qwen2.5 series\footnote{\href{https://huggingface.co/Qwen}{https://huggingface.co/Qwen}}, LLaMA-3.1 series\footnote{\href{https://huggingface.co/meta-llama}{https://huggingface.co/meta-llama}}) and translation-specific (Tower family\footnote{\href{https://huggingface.co/Unbabel/TowerBase-7B-v0.1}{https://huggingface.co/Unbabel/TowerBase-7B-v0.1}}). For each model family, we include both the pre-trained base model and the corresponding instruction-finetuned variant, adapting their chat templates for the Instruct models. 
\begin{tcolorbox}[
    colback=gray!10!white, 
    colframe=darkgreen, 
    boxrule=0.5mm, 
]
\textbf{Finding 4:} LLM architectures exhibit distinct adaptability and effectiveness under MT-R1-Zero, with Qwen showing the highest compatibility in format learning and reasoning generation, while LLaMA and Tower face more challenges and tend towards "format hacking".
\end{tcolorbox}

 As shown in Figure~\ref{fig:train_dynamic_multi}, both the translation-specific (Tower) and LLaMA-3.1 models exhibit significantly slower adaptation to the required \texttt{<think>/<translate>} format compared to Qwen models, as evidenced by their delayed format error reduction. Furthermore, qualitative analysis (Figure~\ref{fig:case_multi_model}) reveals that these models often circumvent meaningful reasoning by generating minimal or templated placeholder content in the \texttt{<think></think>} tags, potentially "hacking" the format reward. In contrast, 
 Qwen2.5 models demonstrate stronger adaptability, consistently producing coherent reasoning text within the structured framework. This suggests that architectures like Qwen may possess inherent advantages for integrating structured reasoning via RL, a finding that aligns with prior work on cognitive behaviors in related domains~\citep{gandhi2025cognitive}. However, even Qwen2.5 models occasionally regress to simplistic one-sentence outputs during reasoning tasks, underscoring the instability of exploration in R1-Zero-like training paradigms.


\begin{table*}[t]
    \centering
    \setlength{\tabcolsep}{3pt}
    \resizebox{\textwidth}{!}{
    \begin{tabular}{lcccccccccc}
        \toprule[0.5mm]
        \multirow{3}{*}{Model} & \multicolumn{4}{c}{In-domain} & \multicolumn{6}{c}{Out-of-distribution} \\
        \cmidrule(lr){2-5} \cmidrule(lr){6-11}
        & \multicolumn{2}{c}{ZH-EN} & \multicolumn{2}{c}{EN-ZH} & \multicolumn{2}{c}{EN-JA} & \multicolumn{2}{c}{DE-ZH} & \multicolumn{2}{c}{DE-EN (Doc)} \\
        \cmidrule(lr){2-3} \cmidrule(lr){4-5} \cmidrule(lr){6-7} \cmidrule(lr){8-9} \cmidrule(lr){10-11}
        & COMETKiwi & XCOMET & COMETKiwi & XCOMET & COMETKiwi & XCOMET & COMETKiwi & XCOMET & COMETKiwi & XCOMET \\
        \midrule

        Qwen2.5-7B (SFT) & 69.29 & 84.80 & 67.25 & 74.29 & 67.77 & 65.39 & 67.01 & 86.17 & 67.44 & 86.74 \\
        Qwen2.5-7B (RL w/o thinking) & 70.78 & 86.26 & 69.62 & 76.03 & 68.68 & 68.77 & 67.84 & 86.67 & 68.31 & 88.30 \\
        Qwen2.5-7B (RL w/ thinking) & 70.81 & 86.17 & 69.43 & 76.36 & 69.27 & 68.49 & 68.74 & 88.69 & 68.74 & 88.69 \\

        \bottomrule[0.5mm]
    \end{tabular}
    }
    \caption{
    Performance comparison of different training paradigms: Supervised Fine-Tuning (SFT) vs. RL with explicit thinking (\textit{RL w/ thinking}) vs. RL without explicit thinking (\textit{RL w/o thinking}). Results shown for in-domain and out-of-distribution tasks support the finding that the RL process itself is the primary driver of gains (Section~\ref{sec:analysis}).
    }
    \label{tab:sft}
\end{table*}

\begin{table}[htbp]
    \centering
    \small
    \resizebox{\columnwidth}{!}{%
    \setlength{\tabcolsep}{5pt}
    \begin{tabular}{@{}lccccc}
        \toprule
        \multirow{2.5}{*}{\sc Model} & \multicolumn{4}{c}{\sc DRT Test Set} \\
        \cmidrule(lr){2-5}
                                      & \sc BLEU & \sc COMETKiwi-22 & \sc XCOMET & Avg. \\
        \midrule
        Qwen2.5-7B-Instruct            & 24.17          & 69.66                & 61.84 & 51.89 \\
        TowerInstruct-13B              & 22.71          & 70.55                & 62.77 & 52.01 \\
        DRT-7B                         & 35.51          & 71.77                & 68.40 & 58.56 \\
        DRT-14B                        & 36.37          & 72.15                & 69.64 & 59.39 \\
        \midrule
        Qwen2.5-7B (SFT)                         & 21.61          & 69.91                & 63.20 & 51.57 \\
        Qwen2.5-7B (RL w/o thinking)              & 28.44          & 72.92                & 66.17 & 55.84 \\
        Qwen2.5-7B (RL w/ thinking)                       & 28.42          & 73.20                & 66.64 & 56.09 \\
        \bottomrule
    \end{tabular}
    }
    \caption{
    Performance comparison on the DRT literature translation dataset~\citep{drt-o1} using BLEU, COMETKiwi-22, and XCOMET metrics.
    }
    \label{tab:drt_results}
\end{table}

\begin{figure}[ht]
    \centering
    \includegraphics[scale=0.3]{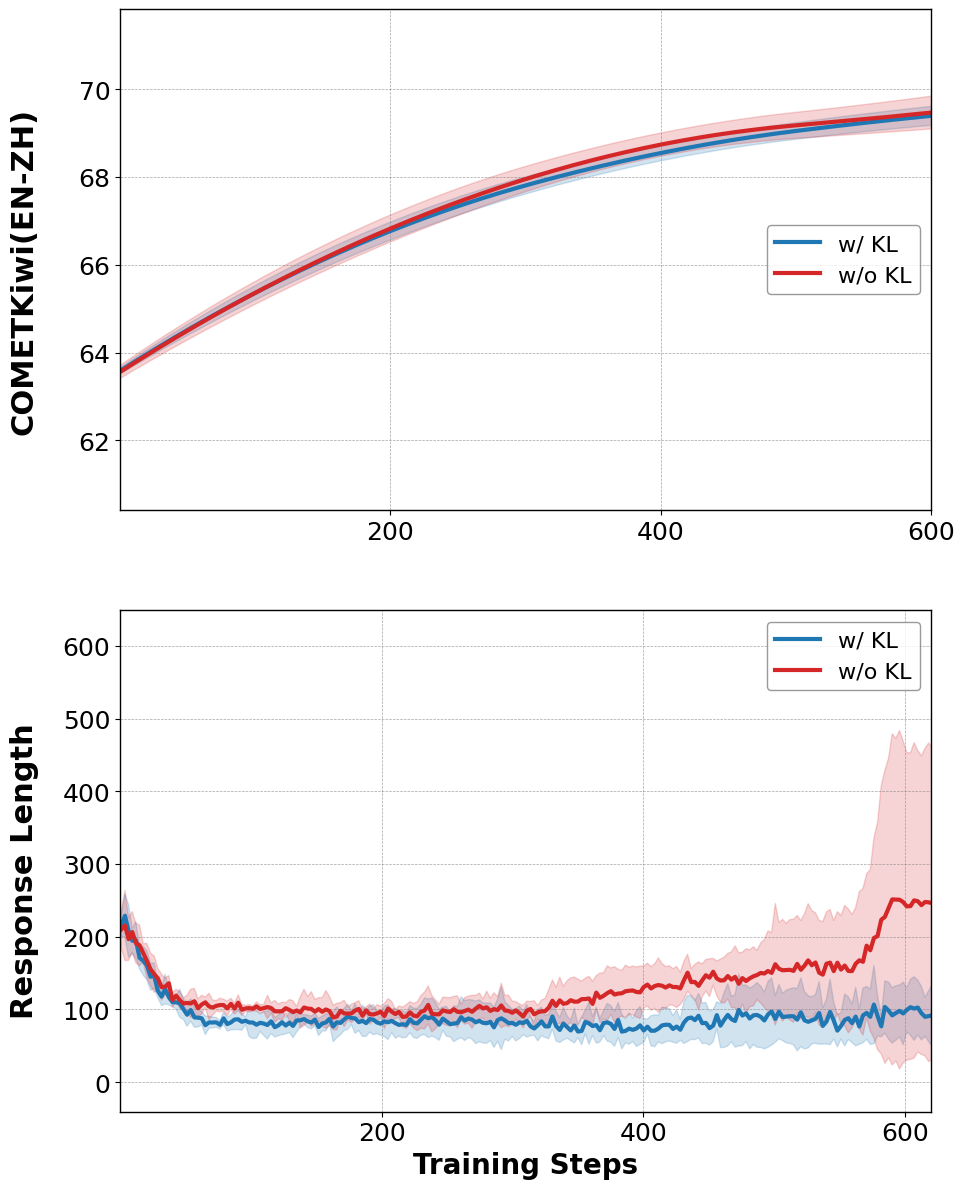}
    \caption{Effect of the KL divergence penalty on EN-ZH COMETKiwi score and response length progression for models trained with (w/ KL, $\beta=0.01$) and without (w/o KL, $\beta=0$) the penalty. Experiments are conducted three times with MT-R1-Zero-7B-Sem.} 
    \vspace{-3mm}
    \label{fig:kl_plot}
\end{figure}

\section{Analysis and Ablation}
\label{sec:analysis}
\subsection{KL Penalty Constrains Response Length but Not Quality Gains}
\label{sec:kl}
We investigate the effectiveness of the KL term in the GRPO objective (Equation~\ref{eq1}) on response length and translation quality, as it would regularize the policy by discouraging large deviations from the initial reference model. We conducted experiments without the KL penalty (setting $\beta=0$, Figure~\ref{fig:kl_plot}), and found that the average response length, after an initial drop, began to fluctuate and trend upward during training. This pattern is consistent with R1-Zero-like results in mathematical tasks~\citep{yu2025dapo,yeo2025demystifying}. Additional ablation of the KL penalty with COMETKiwi reveals that the improvement of translation quality appears to be largely independent of the thinking verbosity. Significant quality gains were achieved in early-stage training (e.g., before Steps 400) before a substantial increase in response length, even in experiments conducted without the KL penalty. This suggests that performance improvements in the MT-R1-Zero setup could not be attributed solely or primarily to increasing reasoning verbosity.

\begin{figure*}[ht]
    \centering
    \includegraphics[scale=0.31]{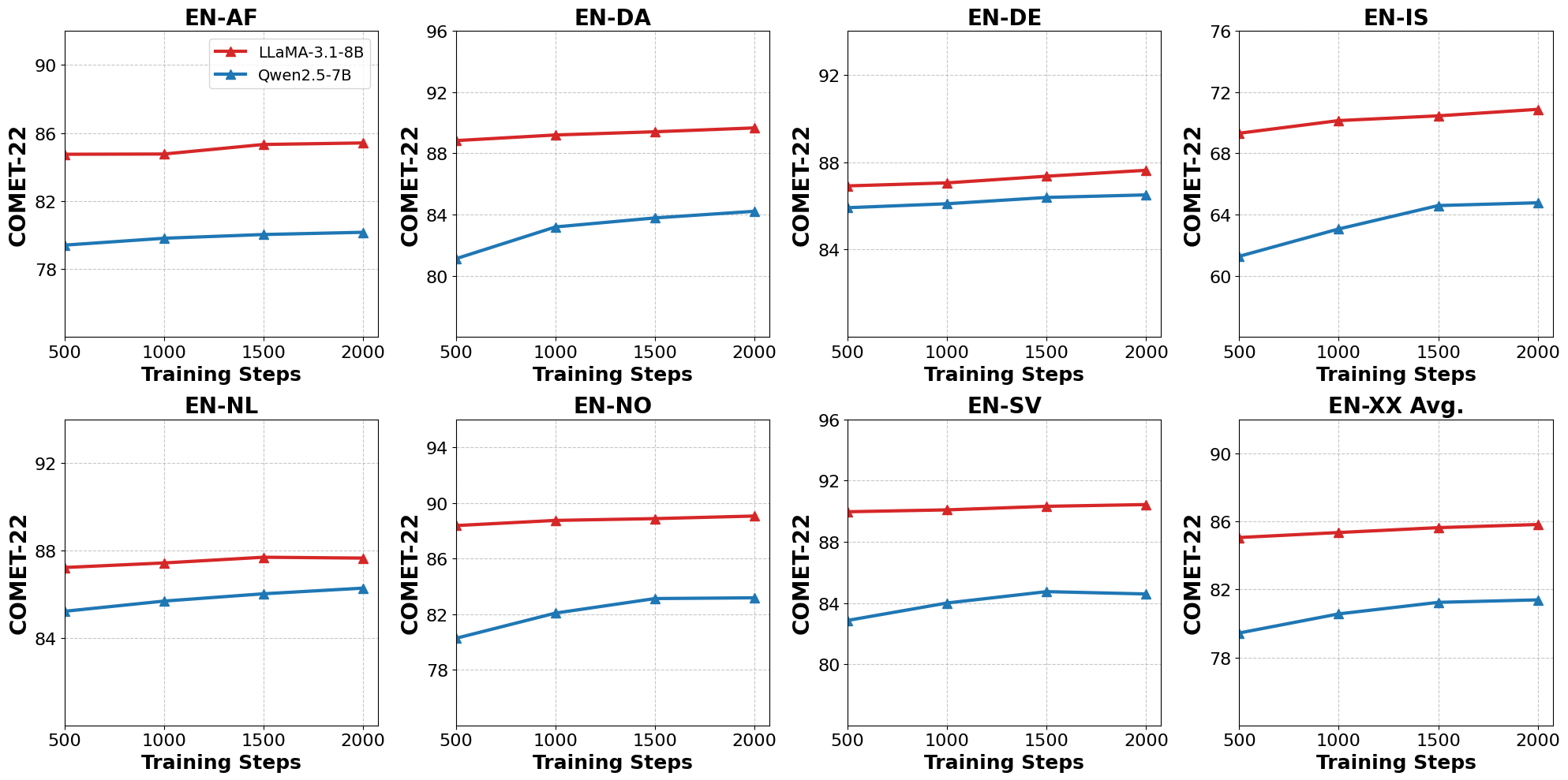}
    \caption{Training progression (COMET-22) for multilingual MT-R1-Zero models based on LLaMA-3.1-8B and Qwen2.5-7B across multiple EN-XX test sets, demonstrating applicability in multilingual settings (Section~\ref{sec:analysis_multilingual}).}
    \vspace{-3mm}
    \label{fig:redstar_comet}
\end{figure*}

\subsection{Disentangling RL and Explicit Thinking}
\label{sec:think}
To determine whether performance gains stem primarily from the explicit \texttt{<think>} step or the underlying RL optimization, we conducted an ablation study comparing three training paradigms using the similar setup from Section~\ref{sec:exp_set}: 1) Supervised Fine-Tuning (SFT): The same base model is fine-tuned on the parallel data using LLaMA-Factory~\citep{zheng2024llamafactory}, establishing a non-RL baseline. 2) RL w/ thinking (MT-R1-Zero-Sem): The model is trained with the rule-metric mixed reward (Format Reward and Reward-Sem) while enforcing explicit \texttt{<think>/<translate>} structure generation. 3) RL w/o thinking: The model is trained with RL-zero optimization (Reward-Sem) solely to the final \texttt{<translate>} output, with no constraints on explicit \texttt{<think>} step generation. See Appendix~\ref{app:sft_details} for more details.

The results are presented in Table~\ref{tab:sft}. It reveals that the "RL w/o thinking" variant achieves performance comparable to MT-R1-Zero ("RL w/ thinking") across both in-domain and OOD tasks, while both RL configurations substantially outperform the SFT baseline – particularly in OOD settings. This pattern is further corroborated by evaluations on the DRT test set (Table~\ref{tab:drt_results}), a literature translation benchmark~\citep{drt-o1}, where we again observe marginal differences between RL variants but significant gains over SFT. These findings demonstrate that while the \texttt{<think>} tag could facilitate emergent reasoning patterns, the major performance improvements in MT-R1-Zero are primarily from the RL framework itself. This aligns with the intuition that online RL methods, iteratively sampling and evaluating self-generated outputs against quality metrics, principally learn "how to translate"  that surpass SFT's behavior cloning limitations.


\subsection{Multilingual and Low-Resource Support}
\label{sec:analysis_multilingual}
To evaluate the broader applicability of our framework, we examine its effectiveness in multilingual training scenarios and its potential benefits for low-resource languages. We train multilingual MT-R1-Zero models using the Germanic language data split in the X-ALMA~\citep{xu2024xalma}, augmented with Chinese (see Table~\ref{tab:multilingual_data} for detailed data statistics). We set the batch size to 16 and used COMET-22\footnote{\href{https://huggingface.co/Unbabel/wmt22-comet-da}{https://huggingface.co/Unbabel/wmt22-comet-da}} as the metric reward (Reward-Sem), consistent with the evaluation protocols in X-ALMA. All models are trained for 1 epoch on 16 NVIDIA H800 80G GPUs for about 12 hours. All other hyperparameters follow the configuration described in Section~\ref{sec:exp_set}.  The training progress, measured by COMET-22 for English-to-target directions, is depicted in Figure~\ref{fig:redstar_comet}. We also report the XCOMET progression in Figure~\ref{fig:redstar_xcomet}. 

The learning curves demonstrate consistent improvement in translation quality across languages spanning diverse resource levels, including those typically considered low-resource (e.g., Icelandic (IS) and Norwegian (NO)). The steady performance improvement observed throughout training confirms that the MT-R1-Zero framework remains effective when applied in multilingual settings.


\section{Conclusion}
In this work, we introduced \textbf{MT-R1-Zero}, the first successful adaptation of R1-Zero RL framework to MT using a novel rule-metric mixed reward mechanism that combines format enforcement with quality metrics. 
Our MT-R1-Zero significantly improves translation quality, achieving leading results on multiple benchmarks, i.e., our 3B models compete with much larger open-source models, while our 7B models are on par with advanced proprietary models. The MT-R1-Zero also demonstrates strong OOD generalization and multilingual applicability. Through extensive experiments and analysis, we highlight the significant impact of reward metric choice for optimization, showcase distinct adaptability across different LLMs, and reveal that performance gains are principally from the RL process itself rather than reasoning steps or verbosity, establishing R1-Zero as a viable and potent paradigm for advancing MT. More broadly, our work highlights the great potential of RL for diverse language processing tasks beyond translation. 

\section*{Limitations}
While MT-R1-Zero represents a significant advance, certain limitations remain. The emergent reasoning observed, though diverse, did not achieve the sophisticated iterative self-correction capabilities demonstrated in mathematical reasoning tasks using similar RL or R1-like methods. This discrepancy may reflect fundamental differences in task structure or indicate the need for specialized design in translation tasks. One promising direction would be developing task-specific cold-start datasets for SFT before RL optimization, though this would deviate from the pure RL paradigm we investigated here. Future work could focus on inducing deeper reasoning structures specifically beneficial for the MT task, investigating architectural adapatability across a broader range of LLMs, and developing more appropriate reward mechanisms. Exploring applications to specialized domains (e.g., law and healthcare) and general language processing tasks presents promising opportunities to extend this work.


\bibliography{custom}

\appendix

\section{Evaluation Details}
\label{app:inference}
When evaluating model performance on the test set,  we deployed open-source models locally using frameworks like vLLM\footnote{\href{https://github.com/vllm-project/vllm}{https://github.com/vllm-project/vllm}}  or HuggingFace\footnote{\href{https://huggingface.co/docs/transformers/main_classes/text_generation}{https://huggingface.co/docs/transformers/\\main\_classes/text\_generation}} implementations.  We use the sampling decoding strategy with a temperature of 0.2, and top\_p set to 0.95. The maximum generation length was capped at 1024 tokens.
We adpot the prompt showcasing in Table~\ref{tab:prompt1} to sample the translation (applying specific chat template when needed).

\section{SFT Training Details}
\label{app:sft_details}
For the Supervised Fine-Tuning (SFT) baseline compared in the ablation study (Section~\ref{sec:think}), we utilized LLaMA-Factory~\citep{zheng2024llamafactory}. The SFT process started from the same base model architecture as the corresponding RL experiments (e.g., Qwen2.5-7B) and was performed on the identical parallel translation dataset (13,130 examples from WMT 2017-2020 after filtering, detailed in Section~\ref{sec:exp_set}). The model was fine-tuned on 8 NVIDIA H800 80G GPUs for 2 epochs using a learning rate of 5e-6 and a batch size of 64, totaling approximately 400 training steps.

\begin{table}
\centering
\resizebox{\columnwidth}{!}{%
    \begin{tabular}{p{\columnwidth}}
    \toprule
    \textbf{Inference Prompt}       \\ 
    \midrule
    Translate the following text from 
    \{src\_language\} into \{tgt\_language\}. \\ 
    \{src\_language\}:\{src\_text\} \\ 
    \{tgt\_language\}: \\
    \bottomrule 
    \end{tabular}
    }
\caption{Prompt used for translation generation. \{tgt\_language\}: target language; \{src\_language\}: source language; \{src\_text\}: the source test sentence.}
\label{tab:prompt1}
\end{table}

\begin{table}[htbp]
    \centering
    \small
    \resizebox{\columnwidth}{!}{%
    \setlength{\tabcolsep}{5pt}
    \begin{tabular}{@{}lcccc}
        \toprule
        \multirow{2.5}{*}{\sc Model} & \multicolumn{4}{c}{\sc Out-of-distribution} \\
        \cmidrule(lr){2-5}
                                      & \sc EN-JA & \sc DE-EN (Doc) & \sc DE-ZH & Avg. \\
        \midrule
        \multicolumn{5}{@{}l}{\textcolor{lightgray}{\textit{Strong Baseline}}} \\
        Qwen2.5-72B-Instruct           & 73.25          & 69.13                & 69.89 & 70.76 \\
        LLaMA3.1-70B-Instruct          & 71.84          & 69.28                & 68.67 & 69.93 \\
        \multicolumn{5}{@{}l}{\textcolor{lightgray}{\textit{Same-size Baseline}}} \\
        Qwen2.5-7B-Instruct            & 64.79          & 67.20                & 67.82 & 66.60 \\
        LLaMA-3.1-8B-Instruct          & 62.42          & 66.77                & 64.28 & 64.49 \\
        TowerInstruct-7B-v0.2          & 58.33          & 69.03                & 65.45 & 64.27 \\
        \midrule
        MT-R1-Zero-7B-Lex                      & 63.33          & 66.17                & 64.32 & 64.61 \\
        MT-R1-Zero-7B-Sem                   & 72.00          & 68.41                & 71.51 & 70.64 \\
        MT-R1-Zero-7B-Mix              & 69.27          & 68.74                & 68.74 & 68.92 \\
        \bottomrule
    \end{tabular}
    }
    \caption{
    Out-of-distribution performance comparison using the COMETKiwi metric on EN-JA, DE-EN (Doc), and DE-ZH. (Complements Table~\ref{tab:ood_xcomet}).
    }
    \label{tab:ood_cometkiwi}
\end{table}

\begin{table}[htbp]
    \centering
    \small
    \resizebox{\columnwidth}{!}{%
    \setlength{\tabcolsep}{5pt}
    \begin{tabular}{@{}lcccc}
        \toprule
        \multirow{2.5}{*}{\sc Model} & \multicolumn{4}{c}{\sc Out-of-distribution} \\
        \cmidrule(lr){2-5}
                                      & \sc EN-JA & \sc DE-EN (Doc) & \sc DE-ZH & Avg. \\
        \midrule
        \multicolumn{5}{@{}l}{\textcolor{lightgray}{\textit{Strong Baseline}}} \\
        Qwen2.5-72B-Instruct           & 25.02          & 45.54                & 40.83 & 37.13 \\
        LLaMA3.1-70B-Instruct          & 24.64          & 45.98                & 37.85 & 36.16 \\
        \multicolumn{5}{@{}l}{\textcolor{lightgray}{\textit{Same-size Baseline}}} \\
        Qwen2.5-7B-Instruct            & 18.91          & 41.17                & 35.25 & 31.78 \\
        LLaMA-3.1-8B-Instruct          & 16.22          & 40.28                & 31.08 & 29.19 \\
        TowerInstruct-7B-v0.2          & 10.52          & 43.40                & 34.74 & 29.55 \\
        \midrule
        MT-R1-Zero-7B-Lex                       & 14.94          & 40.01                & 37.00 & 30.65 \\
        MT-R1-Zero-7B-Sem                       & 14.12          & 33.19                & 22.83 & 23.38 \\
        MT-R1-Zero-7B-Mix                  & 20.27          & 43.17                & 21.41 & 28.28 \\
        \bottomrule
    \end{tabular}
    }
    \caption{
    Out-of-distribution performance comparison using the BLEU metric on EN-JA, DE-EN (Doc), and DE-ZH. (Complements Table~\ref{tab:ood_xcomet}).
    }
    \label{tab:ood_bleu}
\end{table}

\begin{table*}[ht]
    \centering
    \small
    \begin{tabular*}{\textwidth}{@{\extracolsep{\fill}} l ccc ccc ccc @{}}
        \toprule
        & \multicolumn{2}{c}{Train} & \multicolumn{7}{c}{Test} \\
        \cmidrule(lr){2-3} \cmidrule(lr){5-10}
        & EN-ZH & ZH-EN & & EN-ZH & ZH-EN & EN-JA & DE-EN & DE-ZH \\
        \midrule
        \# of cases & 6565 & 6565 & & 997 & 1976 & 997 & 549 & 1012 \\
        Source & \multicolumn{2}{c}{WMT 17-20} & & WMT 24 & WMT 23 & WMT 24 & WMT 23 & Flores \\
        \bottomrule
    \end{tabular*}
    \caption{Data statistics for the training and test sets used in the main experiments (EN$\rightleftharpoons$ZH).}
    \label{tab:main_data}
\end{table*}

\begin{table*}[ht]
    \centering
    \begin{tabular}{lccccc}
        \toprule
        & \multicolumn{4}{c}{Parallel Data} & \\
        \cmidrule{2-6}
         & Train (from EN) & Train (to EN) & Test (from EN) & Test (to EN) & Resource \\
        \midrule
        Afrikaans (AF) & 2994 & 341 & 1012 & 1012 & Mid \\
        Danish (DA) & 2994 & 355 & 1012 & 1012 & Mid \\
        Dutch (NL) & 2994 & 403 & 1012 & 1012 & High \\
        German (DE) & 7015 & 885 & 1012 & 1012 & High \\
        Icelandic (IS) & 4994 & 678 & 1012 & 1012 & Low \\
        Norwegian (NO) & 2994 & 360 & 1012 & 1012 & Low \\
        Swedish (SV) & 2994 & 339 & 1012 & 1012 & High \\
        Chinese (ZH) & 6906 & 874 & 1012 & 1012 & High \\
        English (EN) & - & - & - & - & - \\
        \bottomrule
    \end{tabular}
    \caption{
    Parallel data statistics for languages used in multilingual experiments (Section~\ref{sec:analysis_multilingual}), detailing training/test pairs and resource level classification.
    }
    \label{tab:multilingual_data}
\end{table*}

\begin{figure*}[ht]
    \centering
    \includegraphics[scale=0.31]{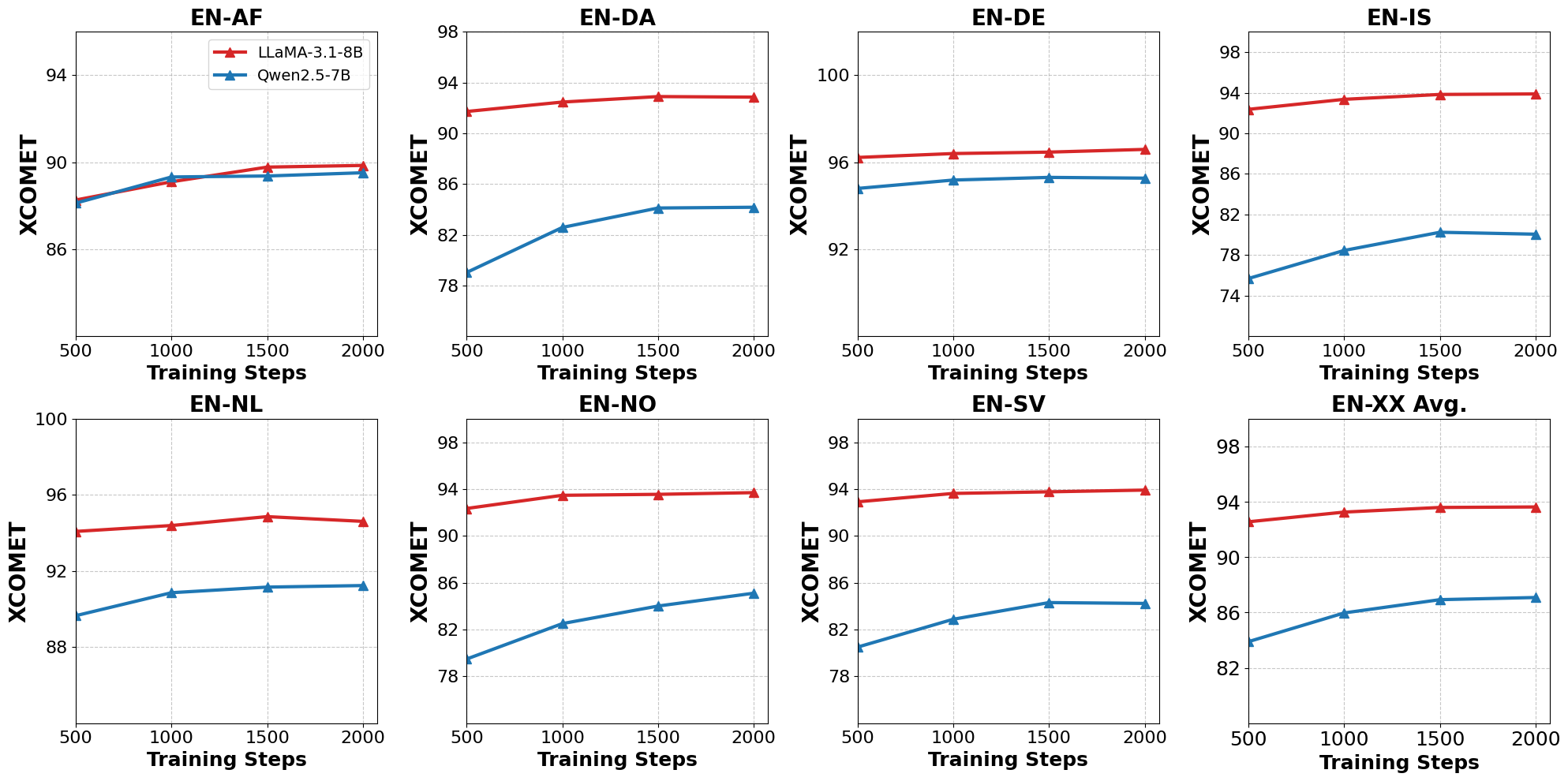}
    \caption{Training progression (reference-free XCOMET score) for multilingual MT-R1-Zero models based on LLaMA-3.1-8B and Qwen2.5-7B across multiple EN-XX test sets (Section~\ref{sec:analysis_multilingual}).} 
    \vspace{-3mm}
    \label{fig:redstar_xcomet}
\end{figure*}

\begin{figure*}[ht]
    \centering
    \includegraphics[scale=0.82]{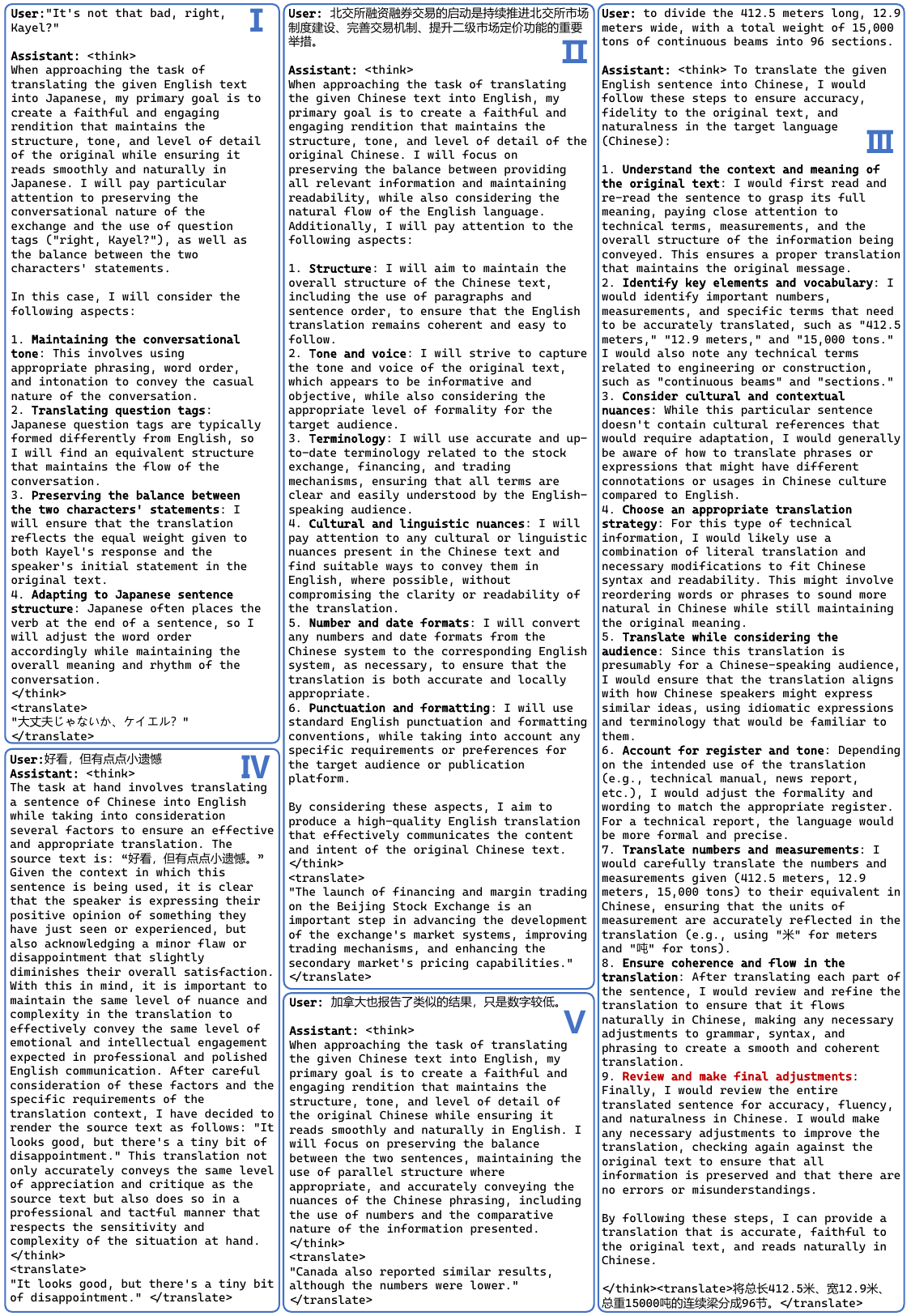}
    \caption{Qualitative examples (I-V) showcasing the diverse thinking patterns generated by MT-R1-Zero models.} 
    \vspace{-3mm}
    \label{fig:case_kl_long_think}
\end{figure*}

\end{document}